# Memory-Efficient Design Strategy for a Parallel Embedded Integral Image Computation Engine

Shoaib Ehsan, Adrian F. Clark, Wah M. Cheung, Arjunsingh M. Bais, Bayar I. Menzat, Nadia Kanwal
and Klaus D. McDonald-Maier
School of Computer Science & Electronic Engineering,
University of Essex,
Colchester, United Kingdom
{sehsan, alien, wmcheu, ambais, bimenz, nkanwa, kdm}@essex.ac.uk

*Abstract*—In embedded vision systems, parallel computation of the integral image presents several design challenges in terms of hardware resources, speed and power consumption. Although recursive equations significantly reduce the number of operations for computing the integral image, the required internal memory becomes prohibitively large for an embedded integral image computation engine for increasing image sizes. With the objective of achieving high-throughput with minimum hardware resources, this paper proposes a memory-efficient design strategy for a parallel embedded integral image computation engine. Results show that the design achieves nearly 35% reduction in memory for common HD video.

## I. INTRODUCTION

Although still relatively new in the image processing domain, the integral image has long been used in computer graphics for texture-mapping as the *summed-area table* [1]. The ground-breaking work of Viola and Jones [2] utilized this intermediate image representation for face detection and paved the way for its use in several multi-scale computer vision algorithms, such as Speeded-Up Robust Features (SURF) [3]. Integral image eliminates time-consuming multiplication operations to allow fast computation of box-type filters in constant time irrespective of their size—a feature which makes it especially attractive for real-time embedded vision applications. The reduction in computational complexity however comes with an associated overhead of calculating this intermediate image representation. Although simple addition operations are required for computing the integral image, the dependence on input image size makes this process computation intensive even for images with medium resolution. For battery-powered embedded vision systems, this presents major challenges in terms of execution speed, hardware resources and power consumption [4]. The focus in this paper is on minimizing the internal memory resources required for an embedded integral image computation engine whilst achieving high throughput. The paper presents a design strategy that achieves nearly 35% reduction in internal memory for common HD video. The remainder of this paper is structured as follows. Section II analyses the internal memory requirements for computing integral image in hardware using recursive equations. Section III proposes a memory-efficient design strategy and presents results for some common image sizes. Finally, conclusions are drawn in Section IV.

## II. PARALLEL COMPUTATION OF INTEGRAL IMAGE USING RECURSIVE EQUATIONS

The value of the integral image corresponding to any location (x,y) in an image may be defined as the sum of all the pixels to the left and above it, including itself. Recursive equations due to Viola and Jones [2] provide a useful method to reduce the total number of addition operations that are required for computation of the integral image. This calculation is serial in nature due to the data dependencies involved. By decomposition of the Viola-Jones recursive equations, it is possible to compute the integral image in row-parallel manner as shown by [4]. However, both the recursion-based serial and parallel methods require one complete row of integral image values to be stored in an internal memory so that it can be utilized for the calculation of the very next row. The width of the required internal memory is log2(number of rows x number of columns x maximum image pixel value) rounded to the upper integer whereas the depth is equal to the total number of columns in one row of the image. Figure 1 highlights the internal memory requirements for an integral image computation engine implemented in hardware for some common images sizes. It is evident that with the increasing image size, the design of the integral image computation engine becomes inefficient in terms of hardware resources due to the large internal memory. It is desirable to investigate a better design which is memory-efficient, with high throughput.

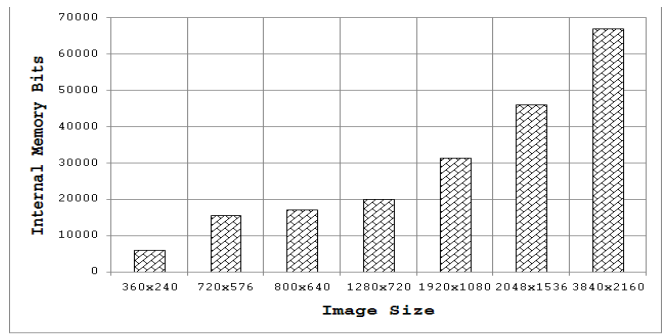

Figure 1. Internal memory requirements for the integral image computation core for some common image sizes.

This research was supported in part by the UK EPSRC Grant EP/I500952/1 and the University of Essex Post-Graduate Research Scholarship awarded to Shoaib Ehsan.

## III. PROPOSED ARCHITECTURE

To address the internal memory problem discussed in Section II, we present a resource-efficient architecture that is also capable of achieving high throughput. The design strategy makes use of the fact that integral image values in adjacent columns of a single row differ by a column sum (please see Figure 2). This difference value is maximum in the last row as the column sum includes all pixel values from the top to the bottom of image in a particular column. In the worst case scenario, the difference between two adjacent columns in the last row of the image will be the product of the number of rows and the maximum value that can be attained by an image pixel (e.g., the maximum value is 255 for an 8-bit pixel).

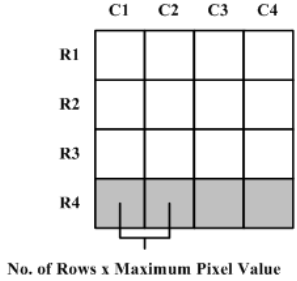

Figure 2. Worst case difference between adjacent integral image values in one row

Figure 3 shows the proposed architecture for an embedded integral image computation engine. This pipelined architecture computes two integral image values in a single clock cycle. Unlike the methods [2, 4] which store a complete row of integral image values in the internal memory for computing the very next row, our design only saves the difference values of the adjacent columns in a row for calculating the next row. Only the integral image value for the first column in that row is saved in a separate register to allow computation of the integral image values from the stored difference values. Although the depth of the internal memory remains the same as mentioned in Section II, the proposed design strategy requires the width to be $\log_2$(number of rows x maximum image pixel value) rounded to the upper integer value. Table 1 provides the results for internal memory reduction when prototyped on an FPGA, a Virtex-6 XC6VLX240T device, for some common image sizes. It is evident from Table 1 that the architecture is capable of achieving significant memory reduction over other recursion-based methods [2, 4] even for small image sizes.

TABLE I. REDUCTION IN INTERNAL MEMORY REQUIREMENTS FOR THE PROPOSED ARCHITECTURE ON VIRTEX-6 XC6VLX240T

| S.No. | Image Size | Internal Memory Bits Required | Reduction w.r.t methods [2, 4] |
|---|---|---|---|
| 1. | 360 x 240 | 4080 | 32% |
| 2. | 720 x 576 | 10368 | 33.3% |
| 3. | 800 x 640 | 11520 | 33.3% |
| 4. | 1280 x 720 | 13680 | 32.1% |
| 5. | 1920 x 1080 | 20520 | 34.4% |
| 6. | 2048 x 1536 | 29184 | 36.6% |
| 7. | 3840 x 2160 | 43200 | 35.4% |

## IV. CONCLUSIONS & FUTURE WORK

This paper has presented a memory-efficient design strategy for a parallel embedded integral image computation engine that is capable of achieving nearly 35% memory reduction for common HD video (1920 x 1080). In future, we intend to design a low-power, high-throughput architecture for SURF detector utilizing the proposed integral image computation engine.

REFERENCES

[1] F. Crow, "Summed-area tables for texture mapping," *ACM SIGGRAPH Computer Graphics,* vol. 18, no. 3, pp. 212, 1984.

[2] P. Viola, and M. Jones, "Rapid Object Detection using a Boosted Cascade of Simple Features," *Proc. IEEE Computer Society Conference on Computer Vision and Pattern Recognition* vol. 1, pp. 511-518, 2001.

[3] H. Bay, A. Ess, T. Tuytelaars, and L. Van Gool, "Speeded-Up Robust Features (SURF)," *Computer Vision and Image Understanding,* vol. 110, no. 3, pp. 346-359, 2008.

[4] S. Ehsan, A. Clark, and K. McDonald-Maier, "Novel Hardware Algorithms for Row-Parallel Integral Image Calculation," 2009 Digital Image Computing: Techniques and Applications, pp. 61-65, Australia, 2009.

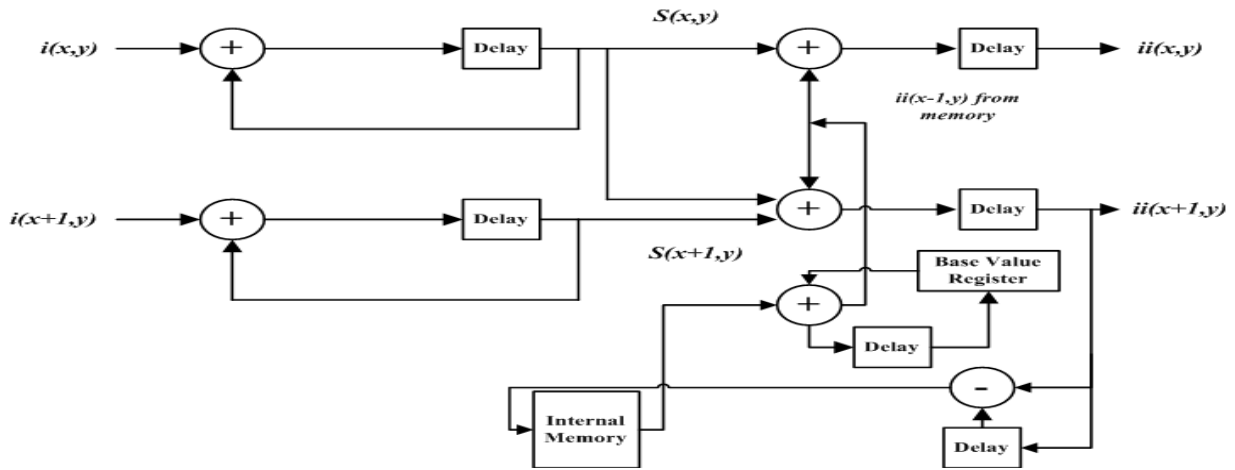

Figure 3. Block diagram of the proposed architecture. i(x,y) and ii(x,y) are the image pixel value and the integral image value at location (x,y) in the image. S(x,y) is the row sum at that particular location.